\documentclass[10pt,twocolumn,letterpaper]{article}

\usepackage[pagenumbers]{cvpr} 

\usepackage{microtype}
\usepackage{graphicx}
\usepackage{booktabs} 

\usepackage{amsmath}
\usepackage{amssymb}
\usepackage{mathtools}
\usepackage{amsthm}

\usepackage[utf8]{inputenc} 
\usepackage[T1]{fontenc}    
\usepackage{url}            
\usepackage{booktabs}       
\usepackage{amsfonts}       
\usepackage{nicefrac}       
\usepackage{microtype}      
\usepackage{xcolor,colortbl}
\definecolor{lgreen}{RGB}{236, 255, 201}
\usepackage{wrapfig}

\usepackage{breqn}
\usepackage{float}
\usepackage{graphicx}
\usepackage{caption}
\usepackage[cal=cm]{mathalfa}
\usepackage{comment}
\usepackage{mathtools}
\usepackage{bbm}

\usepackage{enumitem}
\usepackage{tablefootnote}
\usepackage{footnotehyper}
\usepackage{wrapfig}
\usepackage{floatflt}

\theoremstyle{plain}

\theoremstyle{definition}

\theoremstyle{remark}

\usepackage[ruled]{algorithm}
\usepackage[noend]{algpseudocode}
\usepackage[normalem]{ulem}
\makeatletter

\usepackage[pagebackref=true,breaklinks=true,letterpaper=true,colorlinks,bookmarks=false]{hyperref}

\usepackage[capitalize]{cleveref}
\crefname{section}{Sec.}{Secs.}
\Crefname{section}{Section}{Sections}
\Crefname{table}{Table}{Tables}
\crefname{table}{Tab.}{Tabs.}

\def\cvprPaperID{5314} 
\def\confName{CVPR}
\def\confYear{2023}

\begin{document}


\title{Global Vision Transformer Pruning with Hessian-Aware Saliency}


\author{
Huanrui Yang\textsuperscript{\rm 1,2}\thanks{Work done during an internship at NVIDIA.}, Hongxu Yin\textsuperscript{\rm 1}, Maying Shen\textsuperscript{\rm 1}, Pavlo Molchanov\textsuperscript{\rm 1}, Hai Li\textsuperscript{\rm 3}, and Jan Kautz\textsuperscript{\rm 1}\\ 
\textsuperscript{\rm 1}NVIDIA, \textsuperscript{\rm 2}University of California, Berkeley, \textsuperscript{\rm 3}Duke University\\
{\tt\small huanrui@berkeley.edu, \{dannyy, mshen, pmolchanov, jkautz\}@nvidia.com, hai.li@duke.edu}
}
\maketitle

\begin{abstract}
  Transformers yield state-of-the-art results across many tasks. However, their heuristically designed architecture impose huge computational costs during inference. 
  This work aims on challenging the common design philosophy of the Vision Transformer (ViT) model with uniform dimension across all the stacked blocks in a model stage, where we redistribute the parameters both across transformer blocks and between different structures within the block via the first systematic attempt on \textbf{global} structural pruning. Dealing with diverse ViT structural components, we derive a novel Hessian-based structural pruning criteria comparable across \textit{all} layers and structures, with latency-aware regularization for direct latency reduction.
  Performing iterative pruning on the \textit{DeiT-Base} model leads to a new architecture family called \textit{NViT} (Novel ViT), with a novel parameter redistribution that utilizes parameters more efficiently. 
  On ImageNet-1K, NViT-Base achieves a \textbf{2.6$\times$} FLOPs reduction, \textbf{5.1$\times$} parameter reduction, and \textbf{1.9$\times$} run-time speedup over the \textit{DeiT-Base} model in a near lossless manner. 
  Smaller NViT variants achieve more than \textbf{1\%} accuracy gain at the same throughput of the DeiT Small/Tiny variants, as well as a lossless \textbf{3.3$\times$} parameter reduction over the SWIN-Small model.
  These results outperform prior art by a large margin.
  Further analysis is provided on the parameter redistribution insight of NViT, where we show the \textbf{high prunability} of ViT models, \textbf{distinct sensitivity} within ViT block, and \textbf{unique parameter distribution trend} across stacked ViT blocks. Our insights provide viability for a simple yet effective parameter redistribution rule towards more efficient ViTs for off-the-shelf performance boost.
\end{abstract}
\vspace{-10pt}

\section{Introduction}
Transformer models demonstrate high model capacity, easy scalability, and superior ability in capturing long-range dependency~\cite{vaswani2017attention,devlin2018bert,radford2018improving,jiao2019tinybert,brown2020language}. 
Vision Transformer, \textit{i.e.}, the ViT~\cite{dosovitskiy2020image}, shows that embedding image patches into tokens and passing them through a sequence of transformer blocks can lead to higher accuracy compared to state-of-the-art CNNs. 
DeiT~\cite{touvron2021training} further presents a data-efficient training method such that acceptable accuracy can be achieved without extensive pretraining. 
Offering competitive performance to CNNs under similar training regimes, transformers now point to the appealing perspective of solving both NLP and vision tasks with the same architecture~\cite{zheng2021rethinking,kim2021vilt,jiang2021transgan}.

Unlike CNNs built with convolutional layers that contain few dimensions like the kernel size and the number of filters, the ViT has multiple distinct components, \textit{i.e.}, QKV projection, multi-head attention, multi-layer perceptron, etc.~\cite{vaswani2017attention}, each defined by independent dimensions. As a result, the dimension of each component in each ViT block needs to be carefully designed to achieve a decent trade-off between efficiency and accuracy. However, this is typically not the case for state-of-the-art models. Models such as ViT~\cite{dosovitskiy2020image} and DeiT~\cite{touvron2021training} mainly inherit the design heuristics from NLP tasks, \textit{e.g.,} \textit{use MLP expansion ratio 4}, \textit{fix QKV per head}, \textit{all the blocks having the same dimensions}, etc., which may not be optimal for computer vision~\cite{chen2021autoformer}, causing significant redundancy in the base model and a worse efficiency-accuracy trade-off upon scaling, as extensively shown empirically. 
New developments in ViT architectures incorporate additional design tricks like multi-stage architecture~\cite{wang2021pyramid}, more complicated attention schemes~\cite{liu2021swin}, and additional convolutional layers~\cite{d2021convit} \etc, yet no attempt has been made on understanding the potential of redistributing parameters within the stacked vision transformer blocks.

\begin{figure*}[tb]
\centering
\captionsetup{width=\linewidth}
\includegraphics[width=0.75\linewidth]{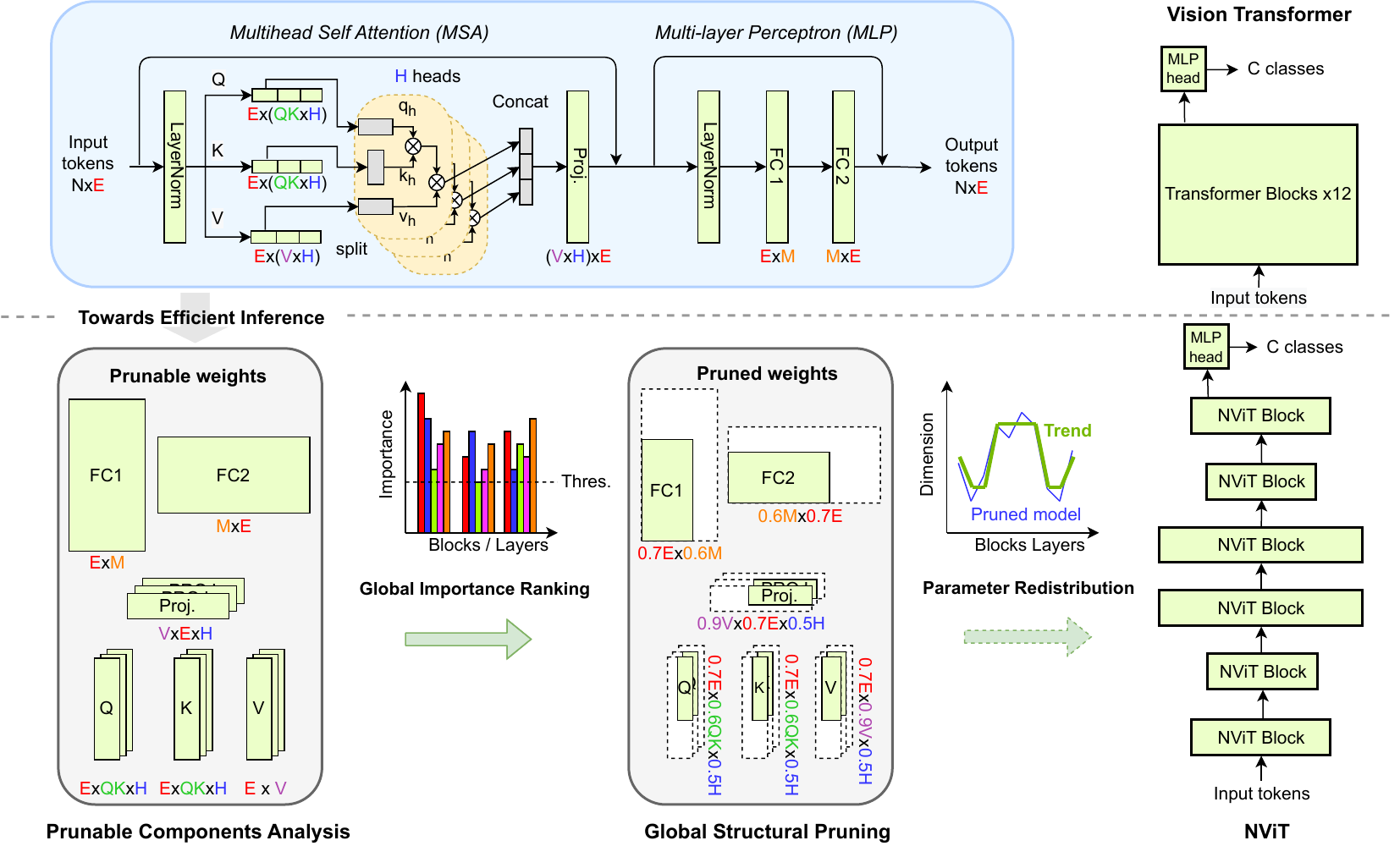}
\caption{\textbf{Towards efficient vision transformer models.} Starting form ViT, specifically DeiT, we identify the design space of pruning (i) embedding size E, (ii) number of head H, (iii) query/key size QK, (iv) value size V and (v) MLP hidden dimension M in~\cref{ssec:dim}. Then we utilize a global ranking of latency-aware importance score to perform iterative global structural pruning in~\cref{ssec:procedure}, achieving pruned NViT models. Finally we analyze the parameter redistribution trend of all the components in the NViT model, as in~\cref{ssec:heur}.}
\label{fig:overall}
\vspace{-15pt}
\end{figure*}

This work targets efficient ViTs by exploring parameter redistribution within ViT blocks and across multiple layers of cascading ViT blocks. To this end, we start with the straightforward DeiT design space, with only ViT blocks. We analyze the importance and redundancy of different components in the DeiT model via latency-aware global structural pruning, leveraging the insights to redistribute parameters for enhanced accuracy-efficiency trade-off. 
Our approach, as visualized in~\cref{fig:overall}, starts from analyzing the blocks in the computation graph of ViT to identify all the dimensions that can be independently controlled. 
We apply global structural pruning over all the components in all blocks. This offers complete flexibility to explore their combinations towards an optimal architecture in a complicated design space. 
Performing global pruning on ViT is significantly challenging, given the diverse structural components and significant magnitude differences. Previous methods only attempts on per-component pruning with the same pruning ratio~\cite{chen2021chasing}, which cannot lead to parameter redistribution across components and blocks.
We derive a new importance score based on the Hessian matrix norm of the loss for global structural pruning, for the first time offering comparability among all prunable components. Furthermore, we 
incorporate the estimated latency reduction into the importance score. This guides the final pruned architecture to be faster on target devices. 

The iterative structural pruning of the DeiT-Base model enables a family of efficient ViT models: NViT.
On the ImageNet-1K benchmark~\cite{ILSVRC15}, NViT enables a nearly lossless 5.14$\times$ parameter reduction, 2.57$\times$ FLOPs reduction and 1.86$\times$ speed up on V100 GPU over the DeiT-Base model. An 1\% and 1.7\% accuracy gain is observed over DeiT-Small and DeiT-Tiny models when we scale down the NViT to a similar latency. NViT achieves a further 1.8$\times$ FLOPs reduction and an 1.5$\times$ speedup over NAS-based AutoFormer~\cite{chen2021autoformer} (ICCV'21) and the SOTA structural pruning method S$^2$ViTE~\cite{chen2021chasing} (NeurIPS'21).
The efficiency and performance benefit of NViT trained on ImageNet also transfers to downstream classification and segmentation tasks. 


Using structural pruning for architectural guidance, we further make an important observation that the popular uniform distribution of parameters across all layers is, in fact, not optimal. To this end, we present further empirical and theoretical analysis on the new parameter distribution rule of efficient ViT architectures, which provides a new angle on understanding the learning dynamic of vision transformer model. We believe our findings would inspire future design of efficient ViT architectures.



Our main contributions are as follows: 
\begin{itemize}
    \item Propose NViT, a novel hardware-friendly \textit{global structural pruning} algorithm enabled by a \textit{latency-aware}, \textit{Hessian-based} importance-based criteria and tailored towards the ViT architecture, achieving a nearly lossless 1.9$\times$ speedup, significantly outperforms SOTA ViT compression methods and efficient ViT designs; 
    \item Provide a systematic analysis on the prunable components in the ViT model. We 
    perform structural pruning on the embedding dimension, number of heads, MLP hidden dimension, QK dimension and V dimension of each head separately; 
    \item Explore hardware-friendly parameter redistribution of ViT, finding \textbf{high prunability} of ViT models, \textbf{distinct sensitivity} within ViT block, and \textbf{unique parameter distribution trend} across stacked ViT blocks.
\end{itemize}

\section{Related work}
\subsection{Vision transformer models}
\label{ssec:vit}





Inspired by the success of transformer models in NLP tasks, recent research proposes to use them on computer vision tasks. The inspiring vision transformer (ViT)~\cite{dosovitskiy2020image} demonstrates the possibility of performing high-accuracy image classification with transformer architecture only. This stimulates recent works to improve training and efficiency of the ViT model. One noticeable approach DeiT~\cite{touvron2021training} provides carefully designed training schemes and data augmentations to train ViT from scratch on ImageNet only. 
Another line of work renovates ViT transformer blocks to better capture image features, such as changing input tokenization~\cite{yuan2021tokens,graham2021levit}, using hierarchical architecture~\cite{liu2021swin,wang2021pyramid,graham2021levit}, upgrading positional encoding~\cite{chu2021conditional}, and performing localized attention~\cite{liu2021swin,han2021transformer}.

In this work we focus on the original ViT architecture~\cite{dosovitskiy2020image} amid its straightforward design space, as illustrated in the top of~\cref{fig:overall}. 
ViT model first divides the input image into patches that are tokenized to embedding dimension $E$ through a linear projection. Image tokens, together with an independently initialized \textit{class token}, form an input $x\in \mathbb{R}^{N\times E}$. Input tokens pass through transformer blocks before classification is made from the class token output of the last block.

A ViT block includes a multi-head self attention (MSA) and a multi-layer perceptron (MLP) module. The MSA module first linearly transforms the $N\times E$ tokens into queries $q\in \mathbb{R}^{N\times (QK\times H)}$, keys $k\in \mathbb{R}^{N\times (QK\times H)}$, and values $v\in \mathbb{R}^{N\times (V\times H)}$. The $q$, $k$ and $v$ are then split into $H$ heads. Each head performs the self-attention operation $\text{Attn}(q_h, k_h, v_h) = \text{softmax}\left(\frac{q_h k_h^T}{\sqrt{d_h}}\right) v_h$ in parallel.
The output of all the heads are then concatenated prior to a fully-connected (FC) linear projection back to the original dimension of $\mathbb{R}^{N\times E}$.
Note that though previous works set $QK = V$ in designing the model architecture~\cite{dosovitskiy2020image,touvron2021training,chen2021autoformer}, setting them differently will not go against the shape rule of matrix multiplication.
The MLP module includes two FC layers with a hidden dimension of $M$. 
The output of the last FC layer preserves token dimension at $\mathbb{R}^{N\times E}$. 

Built upon the original ViT, DeiT models~\cite{touvron2021training} further exploit a \textit{distillation token}, which learns from the output label of a CNN teacher during the training process to incorporate some inductive bias of the CNN model, and significantly improves the DeiT accuracy. 
Our work uses the DeiT model architecture as a starting point, where we explore the potential of better distributing dimensions of different blocks for enhanced efficiency-accuracy tradeoff. 

\subsection{Efficient ViT models}


To improve model efficiency, very recent works perform structural pruning on vision transformer models, with trainable gate variables~\cite{zhu2021visual} or Taylor importance score~\cite{chen2021chasing}. 
Both methods show the potential of compressing ViT models, yet only consider part of the prunable architecture, use \textit{uniform} sparsity for all components, and do not take run time latency into account, thus may not lead to optimal compressed models and cannot discover potential parameter redistribution. Our method resolves these issues through a latency-aware global structural pruning of all prunable components across all layers in a jointly manner.

Besides pruning, multiple attempts have been made in designing efficient ViT architectures. Notable methods include adding convolutional layers~\cite{wu2021cvt,d2021convit}, using multiple ViT stages with different feature scales~\cite{chen2021glit,wang2021pyramid,chen2022auto,zhou2021deepvit}, and explore novel attention mechanisms~\cite{yuan2021tokens,liu2021swin,han2021transformer,hatamizadeh2022global}. Yet all these work use the same dimension for all transformer blocks in each stage, whereas our work explores the parameter redistribution among cascading transformer blocks to achieve better efficiency-accuracy tradeoff without additional tricks. 
The closest attempt to our our work is AutoFormer~\cite{chen2021autoformer}, uses a neural architecture search (NAS) approach to search for parameter redistribution of ViT models. 
Due to the constraint on the supernet training cost, AutoFormer only explores a small number of dimension choices; while our method continuously explores the entire design space of ViT model with a single iterative pruning process, leading to the finding of more efficient architectures.

Another orthogonal yet relevant line of work explores accelerated ViT inference with token pruning~\cite{liang2022evit,rao2021dynamicvit}. Token pruning reduces model FLOPs by halting tokens at early stages without altering the network; while our work removes structural components from weights to reach a smaller \textit{static} architecture. Both ideas are complimentary and we will explore joint pruning in future work.



\section{Latency-aware global structural pruning}

\label{sec:method}

\subsection{Prunable structures with head alignment}
\label{ssec:dim}

\begin{figure*}[htb]
\centering
\captionsetup{width=\linewidth}
\includegraphics[width=0.7\linewidth]{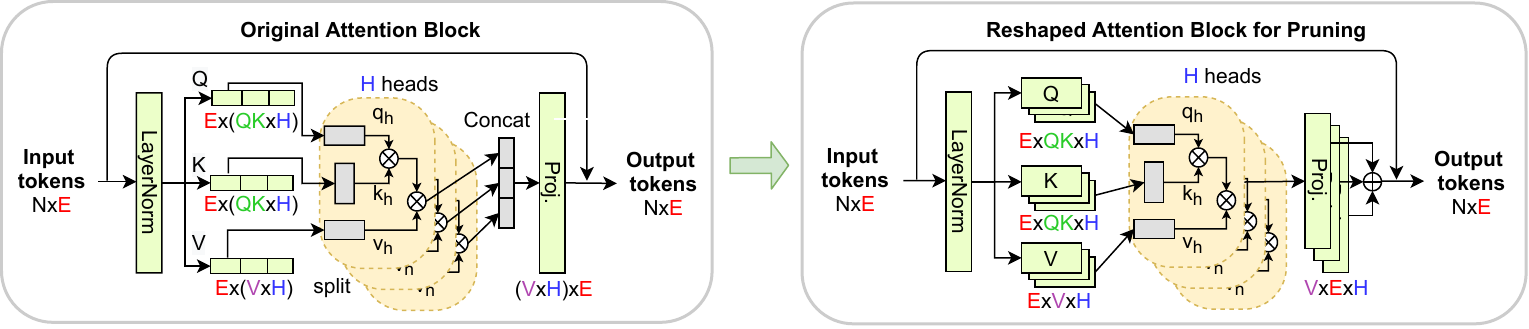}
\caption{\small \textbf{Head Alignment} for latency-friendly pruning. We reshaped the QKV and final output projection in the attention block to explicitly control the number of head and align the QK \& V dimensions in each head.}
\label{fig:prune2}
\vspace{-10pt}
\end{figure*}

To explore the full space of parameter redistribution, we focus on all the independent structures in ViT, namely:
\begin{itemize}[leftmargin=*]
    \item The embedding dimension, denoted as \textit{EMB};
    \item The number of heads in MSA, denoted as \textit{H};
    \item The output dimension of Q and K projection per head in MSA, denoted as \textit{QK};
    \item The output dimension of V projection and input dimension of the PROJ per head, denoted as \textit{V};
    \item The hidden dimension of MLP, denoted as \textit{MLP}.
\end{itemize}

Note that this is slightly different from the dimensions we showed in~\cref{ssec:vit}.
As highlighted on the left of~\cref{fig:prune2}, in a typical ViT implementation, the QKV projection output dimensions are a concatenation of all the attention heads~\cite{rw2019timm}, effectively $QK \times H$ or $V \times H$. The projected tokens are then split into $H$ heads to allow the computation of MSA in parallel. If we directly prune this concatenated dimension, then there is no control on the remaining QK and V dimension of each head. Therefore, the latency of the entire MSA will be bounded by the head with the largest dimension.

To alleviate such inconsistency between pruned head dimensions, we propose \textit{head alignment}, which
explicitly control the number of heads and align the QK and V dimension remaining in each head. As illustrated on the right of~\cref{fig:prune2}, for model pruning we reshape the weight of Q, K, V and PROJ projection layers to single out the head dimension \textit{H}. Performing structural pruning on the reshaped block along the H dimension will enable the removal of an entire head, while pruning along the QK/V dimension guarantees the remained QK and V dimension of all the heads are the same. 
This reshaping is only applied during the pruning process, while the final pruned model is converted back to the concatenated scheme.
Note that H, QK, V and MLP in different blocks can be independently pruned; while EMB needs to be identical across the blocks due to the shortcut connections.

A comparison of pruning with or without head alignment is provided in~\cref{ap:scheme}, where we demonstrate head alignment can bring up to 0.3\% accuracy gain under the same latency target. 

\subsection{Structural pruning algorithm}
\label{ssec:procedure}


\subsubsection{Hessian-based group importance ranking}
\label{sssec:Hessian}
\begin{figure}[h]
\vspace{-10pt}
  \begin{center}
    \includegraphics[width=0.25\textwidth]{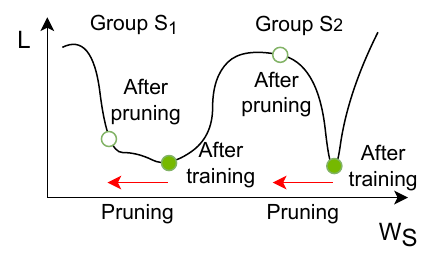}
  \end{center}
  \vspace{-10pt}
  \caption{\textbf{Loss of pruning different structural groups.} Group $\mathcal{S}_1$ with smaller Hessian norm lives in flatter loss minima, leading to lower loss increase after pruning.}
  \label{fig:Hessian}
  \vspace{-10pt}
\end{figure}

Inspired by recent research on the loss surface geometry of deep neural networks, here we consider the Hessian matrix of the loss function with respect to the group of parameters to be pruned to determine our pruning criteria. Specifically, we consider the matrix norm, the squared sum of Hessian eigenvalue, as the criteria for determining the importance of the group of parameters. Previous research~\cite{moosavi2019robustness,yang2021hero,yu2022hessian} has concluded that a smaller Hessian norm indicates a flatter loss surface, which leads to a smaller loss difference when the group is perturbed, i.e. pruned, as in~\cref{fig:Hessian}.

To unify the analysis of structural groups belonging to different components with different shapes and value ranges, we assign a gate variable $g_{\mathcal{S}}$ to each structural group $\mathcal{S}$ of weight, so that the model weight $\mathbf{W}$ is reparameterized as $\mathbf{W} = g_{\mathcal{S}} W_{\mathcal{S}}$, where $W_{\mathcal{S}}$ denotes all weight elements in the structural group $\mathcal{S}$. We set all gates to 1 before pruning so that the reparameterized model is equivalent to the original one. The structural pruning process then aims to find the gates with the smallest Hessian norm, so that we can alter them to 0 to fulfill pruning with minimal loss.

Formally, consider a model whose loss is $\mathcal{L}(\mathcal{D}, g_{\mathcal{S}} W_{\mathcal{S}})$ on dataset $\mathcal{D}$, the Hessian matrix with respect to the gate variables is defined as 
$\mathcal{H}_{i,j} = \frac{\partial^2 \mathcal{L}}{\partial g_{\mathcal{S}_i} \partial g_{\mathcal{S}_j}}$, where $\mathcal{S}_i$ and $\mathcal{S}_j$ are different structural groups. However, a ViT model typically contains tens of thousands of structural groups under our structural pruning configuration, making it infeasible to compute $\mathcal{H}$ directly.
Luckily, here we only need the norm of eigenvalues, i.e. $\sum_i \lambda_i^2$, for our pruning criteria, which can be computed via a Hessian-vector multiplication~\cite{moosavi2019robustness}:
\begin{equation}
\label{equ:Hessian}
    \sum_i \lambda_i^2 = \mathbb{E}_z ||\mathcal{H} z||^2, z\sim \mathcal{N}(0,I).
\end{equation}
With can be further approximated with a finite difference approximation of the Hessian
\begin{equation}
\label{equ:approx}
    \mathcal{H} z \approx (\nabla_{g_{\mathcal{S}}} \mathcal{L}(g_{\mathcal{S}}+hz) - \nabla_{g_{\mathcal{S}}} \mathcal{L}(g_{\mathcal{S}}))/h,
\end{equation}
where $h$ is a small positive constant. This leads to our pruning criteria $\mathcal{I}_{\mathcal{S}}$ as:
\begin{equation}
\label{equ:HesFinal}
    \mathcal{I}_{\mathcal{S}} := \mathbb{E}_z ||(\nabla_{g_{\mathcal{S}}} \mathcal{L}(g_{\mathcal{S}}+hz) - \nabla_{g_{\mathcal{S}}} \mathcal{L}(g_{\mathcal{S}}))/h||^2, z\sim \mathcal{N}(0,1).
\end{equation}
Note that here $z$ follows an univariate normal distribution since $g_{\mathcal{S}}$ is a binary number.

The computation of \cref{equ:HesFinal} is now feasible for all the groups. However, computing the gradient of the gate variable for each group individually is still costly. To efficiently calculate the pruning criteria we simplify~\cref{equ:HesFinal} by further deriving the two gradient terms. Here we derive the second term first since it is simpler. Using the fact $\mathbf{W} = g_{\mathcal{S}} W_{\mathcal{S}}$ and the chain rule we have:
\begin{equation}
\begin{split}
    \label{equ:grad2}
    \nabla_{g_{\mathcal{S}}} \mathcal{L}(g_{\mathcal{S}}) &= \frac{\partial \mathcal{L}}{\partial \mathbf{W}} \frac{\partial \mathbf{W}}{\partial g_{\mathcal{S}}} = (\nabla_{W_{\mathcal{S}}} \mathcal{L}(W_{\mathcal{S}}))^T W_{\mathcal{S}} \\
    &= \sum_{s\in \mathcal{S}} \nabla_{w_s} \mathcal{L}(w_s)\  w_s.
\end{split}
\end{equation}

For the first term, note that by definition $g_{\mathcal{S}}=1$, so $g_{\mathcal{S}}+hz$ is equivalent to $(1+hz)g_{\mathcal{S}}$. In this way we can derive the first term using the result we have in~\cref{equ:grad2} as:
\begin{equation}
\begin{split}
    \label{equ:grad1}
    \nabla_{g_{\mathcal{S}}} \mathcal{L}(g_{\mathcal{S}}+hz) &= \frac{\partial \mathcal{L}}{\partial \mathbf{(1+hz)g_{\mathcal{S}}}} \frac{\partial \mathbf{(1+hz)g_{\mathcal{S}}}}{\partial g_{\mathcal{S}}} \\ 
    &= (1+hz) \sum_{s\in \mathcal{S}} \nabla_{w_s} \mathcal{L}(w_s)\  w_s.
\end{split}
\end{equation}

Substituting~\cref{equ:grad2} and~\cref{equ:grad1} into~\cref{equ:HesFinal} leads to a simplified importance score:
\begin{equation}
\begin{split}
    \label{equ:imp_sim}
    \mathcal{I}_{\mathcal{S}}(\mathbf{W}) &= \mathbb{E}_z || hz \sum_{s\in \mathcal{S}} \nabla_{w_s} \mathcal{L}(w_s)\  w_s /h||^2 \\
    &= \left(\sum_{s\in \mathcal{S}} \mathcal{L}'(w_s)\  w_s\right)^2 \mathbb{E}_z z^2 \\
    &= \left(\sum_{s\in \mathcal{S}} \mathcal{L}'(w_s)\  w_s\right)^2,
\end{split}
\end{equation}
where $\mathcal{L}'(w_s) = \nabla_{w_s} \mathcal{L}(w_s)$.
Since the gradients with respect to all weight elements are already available from backpropagation, the importance score in~\cref{equ:imp_sim} can be easily calculated during the finetuning process without additional cost.
We then greedily remove a few structural groups at a time in our pruning process based on their importance scores, until the targeted constraint is achieved. 

Interestingly, the resulted importance score is similar to the Taylor-based pruning criteria used in CNN filter pruning~\cite{molchanov2019importance,ding2019global,yin2020dreaming,chawla2021data}. Previous work used heuristics to expand the Taylor-based criteria from single parameter importance to structural groups, while we directly derive the structural pruning metric from a novel Hessian-based perspective. 
The Hessian-based importance score can be compared among all layers of weight as a global pruning criteria as it reflects the sensitivity of the structural group to the loss value. Previous pruning methods also considers magnitude-based pruning, which prunes away the group with the lowest weight magnitude. However, we find that magnitude cannot be applied as a global pruning criteria for ViT pruning, as it will make most of the structural components either unpruned or all pruned away. We provide detailed comparison on the effectiveness of our Hessian-based score vs. magnitude-based score for ViT pruning in Appendix~\ref{ap:mag}. 
We also show the strong correlation between our hessian importance score and real loss difference induced by pruning in Appendix~\ref{ap:corr}.


\subsubsection{Latency-aware regularization}

Pruning can be tailored towards latency reduction by penalizing the importance score with latency-aware regularization: 
\begin{equation}
    \label{equ:imp}
    \mathcal{I}^L_{\mathcal{S}}(\mathbf{W}) = \mathcal{I}_{\mathcal{S}}(\mathbf{W}) - \eta \Big( {\text{Lat}}(\mathbf{W})-{\text{Lat}}(\mathbf{W}\backslash \mathcal{S}) \Big).
\end{equation}
$\text{Lat}(\cdot)$ denotes the latency of the current model, which is characterized by a lookup table given the current EMB, H, QK, V, and MLP dimension of each block in the pruned model. Details of the lookup table are provided in Appendix~\ref{ap:lat}, where we show a small lookup table can achieve accurate latency estimation throughout the pruning process.
Latency-aware regularization helps the pruned model to reach the latency target faster with higher accuracy, as shown in Appendix~\ref{ap:latreg}.
We use $\mathcal{I}^L_{\mathcal{S}}$ as the pruning criteria for iterative pruning in our work, with detailed procedure in Appendix~\ref{ap:config}. 
A compact and dense model can be achieved by removing pruned groups and recompiling the model.

\subsubsection{Ampere (2:4) GPU sparsity}


The recently introduced NVIDIA Ampere GPU supports acceleration of sparse matrix multiplication with a specific pattern of 2:4 sparsity (2 of the 4 consecutive weight elements are zero). This comes with a limitation of requiring the input and output dimensions of all linear projections to be \textbf{divisible by 16}~\cite{mishra2021accelerating}. We assure compatibility with such pattern by structurally pruning matrices to have the remaining dimension be divisible by 16 (more details in Appendix~\ref{ap:config}). Interestingly, we find that Ampere sparsity can be performed \textbf{losslessly} with magnitude pruning after the initial pruning.   


\subsection{Training objective}
\label{ssec:objective}

We next consider the training objective function that supports both pruning for importance ranking and finetuning for weight update. To start with, we inherit the CNN hard distillation training objective as proposed in DeiT~\cite{touvron2021training}, which is formulated as follows:
\begin{equation}
    \label{equ:CNN}
    \mathcal{L}_{\text{CNN}} =\mathcal{L}_{\text{CE}} \Big( \Psi(z^s_c),Y \Big)+ \mathcal{L}_{\text{CE}} \Big(\Psi(z^s_d),Y^{\text{CNN}} \Big),
\end{equation}
where $\Psi(\cdot)$ denotes softmax and $\mathcal{L}_{\text{CE}}$ the cross entropy loss. We refer to logits computed from the \textit{class token} of the pruned model as $z_c^s$, and the one computed from the \textit{distillation token} as $z_d^s$. Note that $z_c^s$ is supervised by the true label $Y$, while $z_d^s$ is supervised by the output label of a CNN teacher $Y^{\text{CNN}}$. Unless otherwise stated, we use a pretrained RegNetY-16GF model~\cite{radosavovic2020designing} as the teacher, in line with DeiT. 

In addition to CNN distillation, we consider \textit{full model distillation} given the unique access to such supervision under the pruning setup. Specifically, the ``full model'' corresponds to the pretrained model, which serves as the starting point of the pruning process. 
Ideally a pruned model shall behave similar to its original counterpart. To encourage this, we distill the classification logits from both the class and distillation tokens of the pruned model from the original counterpart, forming~\cref{equ:full}:
\begin{equation}
    \label{equ:full}
    \mathcal{L}_{\text{full}} = \mathcal{L}_{\text{KL}} \Big(\Psi(z^s_c / \tau),\Psi(z^t_c / \tau)\Big)+ \mathcal{L}_{\text{KL}}\Big(\Psi(z^s_d / \tau),\Psi(z^t_d / \tau)\Big).
\end{equation}
Superscripts $^t$ and $^s$ denote the output of the pretrained model and the model being pruned respectively. $\mathcal{L}_{\text{KL}}$ is the KL divergence loss, and $\tau$ is the distillation temperature.

The final objective is therefore composed as: $\mathcal{L} = \alpha \mathcal{L}_{\text{full}} + \mathcal{L}_{\text{CNN}}$. 
An ablation study of alternating the formulation of the training objective is provided in Appendix~\ref{ap:obj}.


\section{NViT Performance}

\subsection{Pruning analysis on ImageNet-1K}
\label{ssec:pruning_results}
We apply our pruning method on the challenging ImageNet-1K benchmark, using the DeiT-Base model pretrained with CNN distillation as the starting point to achieve a family of NViT models. The training and finetuning hyperparameters can be found in Appendix~\ref{ap:hyper}.

\textbf{Comparing with existing models.}
We compare the model size, run time speedup and accuracy of the state-of-the-art manually designed ViT models and our pruned models in Table~\ref{tab:full_compress}. 
For best insights, we conduct pruning in four configurations. Note that all these 4 configuration are achieved from the same pretrained DeiT-Base model in a single global pruning run, each finetuned from a checkpoint snapshot after different pruning steps. Details for our pruning configurations can be found in Appendix~\ref{ap:config}. 
\begin{itemize}[leftmargin=*]
    \item \textbf{NViT-B} aims to match the accuracy of DeiT-B model, which achieves an 1.86$\times$ speedup and a 2.57$\times$ reduction on FLOPs over DeiT-B with neglectable 0.07\% accuracy drop. It also achieves a lossless 2.25$\times$ further FLOPs reduction over the more efficient SWIN-B model.
    \item \textbf{NViT-H} aims to half the latency of DeiT-B, with only 0.4\% accuracy loss. It also achieves 1.41$\times$ further reduction on FLOPs over SWIN-S with similar accuracy.
    \item \textbf{NViT-S} matches DeiT-S latency, with $+1 \%$ accuracy.
    \item \textbf{NViT-T} matches DeiT-T latency, with $+1.7 \%$ accuracy.
\end{itemize}

Furthermore, the superiority of NViT over DeiT and SWIN \textbf{cannot be bridged} even after we finetune the pretrained models. For example, finetuning the pretrained DeiT-T, DeiT-S, and SWIN-T models for additional 300 epochs following the scheme of NViT finetuning will improve the accuracy to 75.0\%, 81.8\%, and 81.7\% respectively, which are still below what achieved by the corresponding NViT models. The lossless 1.9x model acceleration for DeiT-B with the NViT-B configuration has never been achieved from previous designs.

\begin{table}[tb]
  \caption{\small \textbf{Structural pruning results on ImageNet-1K.} Our NViT models are compared with manually designed ViT architectures. All compression ratios and speedups are computed with respect to that of DEIT-Base model. All Latency estimated on a single GPU with batch size 256. ``ASP'' means post-training 2:4 Ampere sparsity pruning with TensorRT~\cite{mishra2021accelerating}.
  }
  \label{tab:full_compress}
  \centering
  \resizebox{0.9\linewidth}{!}{
  \begingroup
    \setlength{\tabcolsep}{3pt}
  \begin{tabular}{c|ccccc}
    \toprule
      &   \multicolumn{2}{c}{Size (Compression)} & \multicolumn{2}{c}{Speedup ($\times$)} &  \\
    \cmidrule(l{2pt}r{2pt}){2-3}
    \cmidrule(l{2pt}r{2pt}){4-5}
    Model & \#Para ($\times$)   & \#FLOPs ($\times$) & V100 & RTX 3080 & Top-1 Acc.     \\
    \midrule
    DEIT-B   & 86M (1.00)     & 17.6G (1.00) & 1.00 & 1.00 &  83.36      \\
    SWIN-B   & 88M (0.99)     & 15.4G (1.14)  & 0.95 & - & 83.30             \\
    \rowcolor{lgreen}
    \textbf{NViT-B}   & 34M (2.57)   & 6.8G (2.57) & \textbf{1.86} & 1.75 & 83.29    \\
    \rowcolor{lgreen}
    \textbf{ + ASP}   & 17M (5.14)   & 6.8G (2.57) & \textbf{1.86} & \textbf{1.85} & 83.29    \\
    \midrule
    SWIN-S   & 50M (1.74)      & 8.7G (2.02) & 1.49 & - & 83.00             \\
   \rowcolor{lgreen}
   \textbf{NViT-H}   & 30M (2.84)   & 6.2G (2.85) & \textbf{2.01} & 1.89 & 82.95     \\
   \rowcolor{lgreen}
   \textbf{ + ASP}   & 15M (5.68)   & 6.2G (2.85) & \textbf{2.01} & \textbf{1.99} & 82.95     \\
    \midrule
    DEIT-S  & 22M (3.94)      & 4.6G (3.82)  & 2.44  &  2.27 & 81.20     \\
    SWIN-T  & 29M (2.99)      & 4.5G (3.91)  & 2.58 & - & 81.30       \\
    \rowcolor{lgreen}
    \textbf{NViT-S}   & 21M (4.18) & 4.2G (4.24) & \textbf{2.52} & 2.35 & \textbf{82.19}    \\
    \rowcolor{lgreen}
    \textbf{ + ASP}   & 10.5M (8.36) & 4.2G (4.24) & \textbf{2.52} & \textbf{2.47} & \textbf{82.19}    \\
    \midrule
    DEIT-T   & 5.6M (15.28)  & 1.2G (14.01) & 5.18 & 4.66 & 74.50     \\
    \rowcolor{lgreen}
    \textbf{NViT-T}   & 6.9M (12.47)    & 1.3G (13.55) & 4.97  & 4.55 & \textbf{76.21}     \\
    \rowcolor{lgreen}
    \textbf{ + ASP}   & 3.5M (24.94)    & 1.3G (13.55) & 4.97 & \textbf{4.66} & \textbf{76.21}     \\
    \bottomrule
  \end{tabular}
  \endgroup
}

\end{table}

\textbf{Comparing with SOTA compression methods.}
We compare NViT with state-of-the-art ViT compression methods, AutoFormer~\cite{chen2021autoformer} in ICCV'21, S$^2$ViTE~\cite{chen2021chasing} in NeurIPS'21, EViT~\cite{liang2022evit} in ICLR'22, and SPViT~\cite{he2021Pruning} in Table~\ref{tab:SOTA}. 
For a fair comparison for all methods we report the accuracy trained with CNN hard distillation. As no such accuracy is available in the S$^2$ViTE paper, we rerun the experiment with CNN distillation following their official GitHub repo\footnote{\url{https://github.com/VITA-Group/SViTE}}.
\begin{itemize}[leftmargin=*]
    \item \textbf{Comparing to AutoFormer}: NViT-H achieves a further 1.5$\times$ speedup over AutoFormer-B with a higher accuracy; NViT-T outperforms AutoFormer-T by 0.5\% under similar size and lower latency. 
    \item \textbf{Comparing to S$^2$ViTE}: NViT-H achieves a further 1.9$\times$ FLOPs reduction and 1.5$\times$ speedup over the 40\%-pruned model, with a higher accuracy.
    \item \textbf{Comparing to EViT}: NViT-S achieves a further 2.8$\times$ FLOPs reduction and 1.6$\times$ speedup over the pruned Base model, with a higher accuracy.
\end{itemize}
Moreover, the lossless 1.9$\times$ speedup of NViT-B over DeiT-B is a big leap over all previous methods.

\begin{table}[tb]
  \caption{\small \textbf{Comparing with SOTA ViT efficiency improvement methods.} S$^2$ViTE and EViT speedups are taken from their papers, while AutoFormer speedup is measured with the same code base as NViT on a RTX 3080 GPU. All speedups are computed with respect to that of DeiT-Base model.}
  \label{tab:SOTA}
  \centering
  \resizebox{0.35\textwidth}{!}{
  \begingroup
    \setlength{\tabcolsep}{4pt}
  \begin{tabular}{c|ccc}
    \toprule
    Model & \#FLOPs & Speedup & Top-1 Acc.      \\
    \midrule
    \rowcolor{lgreen}
    \textbf{NViT-B}   & \textbf{6.8G} & \textbf{1.85$\times$} & \textbf{83.29}    \\
    \midrule
    S$^2$ViTE-B-40~\cite{chen2021chasing} & 11.7G & 1.33$\times$ & 82.92 \\
    AutoFormer-B~\cite{chen2021autoformer} & 11G & 1.34$\times$ & 82.90             \\
    SPViT~\cite{he2021Pruning} & 8.4G & - & 82.40 \\
   \rowcolor{lgreen}
   \textbf{NViT-H}  & \textbf{6.2G} & \textbf{1.99$\times$} & \textbf{82.95}     \\
    \midrule
    EViT-DeiT-B~\cite{liang2022evit} & 11.6G & 1.59$\times$ &  82.10      \\
    \rowcolor{lgreen}
    \textbf{NViT-S} & \textbf{4.2G} & \textbf{2.47$\times$} & \textbf{82.19} \\
    \midrule
    AutoFormer-T~\cite{chen2021autoformer}    & 1.3G & 4.59$\times$ & 75.70   \\
    \rowcolor{lgreen}
    \textbf{NViT-T}   & 1.3G & \textbf{4.66$\times$} & \textbf{76.21}     \\
    \bottomrule
  \end{tabular}
  \endgroup
}
\vspace{-5pt}
\end{table}


\textbf{Comparing with concurrent ViT variants.}
NViT provides a viable way to discover efficient architecture with parameter redistribution in the DeiT design space, without using additional components like more layers, specially designed attention, or multi-stage architecture. Here we compare NViT with concurrent ViT architectures in~\cref{tab:asvit}.
NViT models achieve stronger performance than these architectures while only exploring the basic DeiT design space.

\begin{table}[tb]
  \caption{\small \textbf{Comparing with concurrent ViT architectures.} Accuracy with or w/o CNN distillation are reported when available.}
  \label{tab:asvit}
  \centering
  \resizebox{0.4\textwidth}{!}{
  \begingroup
    \setlength{\tabcolsep}{2pt}
  \begin{tabular}{c|cccc}
    \toprule
    Model & \#Para & \#FLOPs & Acc. (no dis)  & Acc. (dis)    \\
    \midrule
    ConViT-S+~\cite{d2021convit} & 48M & 10G & 82.2 & 82.9 \\
    CaiT-S-24~\cite{touvron2021going} & 46.9M & 9.4G & 82.7 & 83.5 \\
    CaiT-XS-36~\cite{touvron2021going} & 38.6M & 8.1G & 82.6 & 82.9 \\
    \rowcolor{lgreen}
    \textbf{NViT-B}   & \textbf{34M} & \textbf{6.8G} & \textbf{82.8} & \textbf{83.3}    \\
    \midrule
    T2T-ViT-14~\cite{yuan2021tokens} & 21.5M & 6.1G & 81.7 & - \\
    CaiT-XS-24~\cite{touvron2021going} & 26.6M & 5.4G & 81.8 & 82.0 \\
    As-ViT-S~\cite{chen2022auto} & 29.0M & 5.3G & 81.2 & - \\
    TNT-S~\cite{han2021transformer} & 23.8M & 5.2G & 81.5 & - \\
    CvT-13~\cite{wu2021cvt} & 20M & 4.5G & 81.6 & - \\
    GLiT-S~\cite{chen2021glit} & 24.6M & 4.4G & 80.5 & - \\
    PVT-S~\cite{wang2021pyramid} & 24.5M & 3.8G & 79.8 & - \\
    \rowcolor{lgreen}
    \textbf{NViT-S} & \textbf{21M} & \textbf{4.2G} & \textbf{82.0} & \textbf{82.2} \\
    \bottomrule
  \end{tabular}
  \endgroup
}
\vspace{-10pt}
\end{table}



\textbf{Pruning other ViT variants.} We try NViT on pruning the SWIN transformer model. Note that SWIN transformer doesn’t bring additional structural components comparing to ViT, as the novel shift-window attention mechanism is parameter-free. In this case our method can be applied on a single stage in SWIN-Transformer directly without any modification. Here we prune stage 2 of the SWIN-B model, which consists of 18/24 of the transformer blocks, 65\% of parameters, 75\% of FLOPs and 70\% of the overall latency. NViT achieves a \textbf{\textit{lossless}} Stage 2 compression of $1.8\times$ parameter reduction, $1.8\times$ FLOPs reduction and $1.7\times$ runtime speedup on V100 GPU.This indicates that NViT is also applicable to other ViT variants. 

\subsection{Transfer learning to downstream tasks}

Finally, we evaluate the generalizability of our pruned NViT models. Here we finetune the ImageNet trained DeiT and NViT models on CIFAR-10, CIFAR-100~\cite{krizhevsky2009learning}, iNaturalist 2018 and 2019~\cite{van2018inaturalist} dataset. We further investigate the potential of transferring the achieved NViT models into backbones for tasks beyond classification, specifically, semantic segmentation. We evaluate the performance of DeiT/NViT backbones on the Cityscape dataset~\cite{Cordts2016Cityscapes} and the ADE20K dataset~\cite{zhou2017scene}.
The details of the datasets used for our transfer learning experiments and detailed experiment settings are provided in Appendix~\ref{ap:downstream}.
Results are provided in~\cref{tab:downstream}. NViT models consistently outperform the DeiT models on all the tasks. These observations show that the efficiency demonstrated on ImageNet can be preserved on downstream tasks, even beyond classification.

\begin{table}[tb]
  \caption{\small \textbf{Transfer learning tasks performance with ImageNet pretraining.} We report the performance of finetuning the ImageNet trained models on other datasets. Top-1 accuracy is reported for classification tasks, while mIoU is reported for segmentation tasks}
  \label{tab:downstream}
  \centering
  \resizebox{\linewidth}{!}{%
  \begingroup
    \setlength{\tabcolsep}{4pt}
  \begin{tabular}{c|cccc|cc}
    \toprule
    Model & CIFAR-10 &  CIFAR-100 &  iNat-18 & iNat-19 & Cityscape & ADE20K \\
    \midrule
    DeiT-S & 98.52\% & 87.07\% & 66.79\% & 74.22\% & 71.89\% & 40.15\%  \\
    \rowcolor{lgreen}
    \textbf{NViT-S}  & \textbf{98.78\%} & \textbf{87.90\%} & \textbf{69.10\%} & \textbf{77.00\%} & \textbf{73.22\%} & \textbf{41.54\%} \\
    \midrule
    DeiT-T & 97.93\% & 85.66\% & 62.41\% & 72.08\% & 66.65\% & 34.38\% \\
    \rowcolor{lgreen}
    \textbf{NViT-T}  & \textbf{98.31\%} & \textbf{85.88\%} & \textbf{64.78\%} & \textbf{74.65\%} & \textbf{67.09\%} & \textbf{35.42\%}  \\
    \bottomrule
  \end{tabular}%
  \endgroup
  }
  \vspace{-10pt}
\end{table}

\section{Exploring parameter redistribution}
\label{sec:result}
\subsection{Trends observed in ViT pruning}
\label{ssec:heur}
\begin{figure*}[htb]
\centering
\captionsetup{width=\linewidth}
\includegraphics[width=0.8\linewidth]{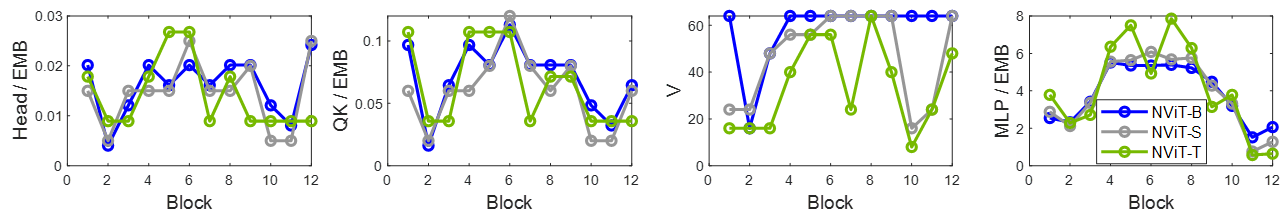}
\caption{\small Model dimension comparison between NViT-B (blue), NViT-S (grey) and NViT-T model (green).}
\label{fig:heur}
 \vspace{-10pt}
\end{figure*}

As observed by~\cite{liu2018rethinking}, channel/filter pruning in CNN models can provide guidance on finding efficient network architectures, yet this has never been explored on ViT models. Here we show \textit{for the first time} that our pruning method can serve as an effective architecture search tool for ViT models.
We observe NViT models of different sizes follows consistent insights, as visualized in~\cref{fig:heur}:
\begin{enumerate}[leftmargin=*]
    \item Number of heads, QK of each head and MLP scales \textit{linearly} with the dimension of EMB; while V of each head can be largely kept the same;
    \item \textit{Reducing} dimensions related to the multi-head attention (H, QK) while \textit{increasing} MLP dimension may lead to more accurate model under similar latency.
    \item The scaling factors of head, QK and MLP are \textit{not uniform} among all blocks: dimension is larger in the blocks in the middle and smaller towards the two ends;
\end{enumerate}

Compared to original ViT design, our insight shows, within each block, the need to scale QK separately from V, and more importantly to distribute different dimensions across different ViT blocks. Interestingly, these trends are not observed in NLP transformer compression~\cite{michel2019sixteen,voita2019analyzing}. 


\subsection{Understanding the parameter redistribution}
\label{ssec:analysis}

\begin{table}[tb]
  \caption{\small \textbf{Average Hessian trace and latency} (V100, batch size 576) per neuron in each structure of the DeiT-B model.}
  \label{tab:hess}
  \centering
  \resizebox{\linewidth}{!}{
  \begin{tabular}{c|cccccc}
    \toprule
    Structure  & Q   & K    & V  & Proj   &   FC1 & FC2 \\
    \midrule
    Hessian trace   & 1.4e-6 & 1.6e-6 & 6.5e-6 & 6.4e-6 & 6.1e-6 & 4.6e-6 \\
    Latency (s)     & \multicolumn{2}{c}{1.7e-4} & \multicolumn{2}{c}{1.4e-4} & \multicolumn{2}{c}{1.2e-5}  \\
    \bottomrule
  \end{tabular}
  }
\vspace{-5pt}
\end{table}

Given the insights on the parameter redistribution trend, here we analyze its reason from the perspective of Hessian sensitivity analysis. Averaged Hessian trace of the training loss with respect to the model weights has been shown effective for analyzing the importance of different structural components in a DNN model~\cite{dong2019hawq,yu2022hessian}. Here we compute the per-structure average Hessian trace of the DeiT-B model on ImageNet in~\cref{tab:hess}. Average latency reduction in pruning each neuron is also provided. Comparing across different structures, we can see V/Proj appears more important than Q/K, showing the need to scale them separately (insight 1). MLP layers also show higher importance than QK layers, while occupying less latency. This justifies redistributing parameters from QK to MLP layers for better latency-accuracy tradeoff (insight 2). 
Besides Hessian, Appendix~\ref{ap:intuition} observes the trend in the attention score diversity among all heads of each block, which reflects a similar less-more-less trend in redundancy appears in each block (insight 3).

\subsection{Comparing to CNN}

As global structural pruning has been extensively studied on CNN, here we compare our insights achieved in NViT with the results in SOTA CNN pruning research~\cite{molchanov2019importance,shen2021halp}:
\begin{itemize}
    \item \textbf{Prunability:} ViT appears to have \textbf{\textit{higher prunability}} than CNN models. SOTA CNN pruning achieves lossless 2$\times$ FLOPs reduction and 1.6$\times$ speedup on ResNet models~\cite{shen2021halp}. Whereas we achieve lossless 2.6$\times$ FLOPs reduction and 1.9$\times$ speedup on DeiT-B model;
    \item \textbf{Structure diversity:} Convolutional layers within a CNN block typically show comparable sensitivity~\cite{molchanov2019importance}. Whereas different structural components within a ViT block shows \textbf{\textit{distinct sensitivity}} in pruning. 
    \item \textbf{Sensitivity distribution:} Sensitivity is lower in the earlier layers of a CNN stage, then gradually increase towards the end~\cite{molchanov2019importance}. Whereas we discovered a unique \textbf{\textit{less-more-less}} distribution among stacked ViT blocks.
\end{itemize}
These comparisons show the different challenges and opportunities faced by efficient CNN and ViT designs.
We hope our study can inspire future exploration on the different learning dynamics and architecture design rules between CNN and ViT architectures.

\subsection{Design novel architecture with redistribution}
\label{ssec:nvit}

\textbf{ViT parameter redistribution.} To further illustrate the effectiveness on our insights on the redistribution of parameters, we follow our insights to design a new architecture we name \textit{ReViT} (Redistributed ViT). We follow the trends in~\cref{fig:heur} and heuristically design a simplified rule in~\cref{tab:heur_sim} to determine the parameter dimensions of each block. 
\begin{table}[tb]
  \caption{\small \textbf{ReViT block dimensions.} For comparison the dimensions of a DeiT block are also listed.}
  \label{tab:heur_sim}
  \centering
  \resizebox{0.8\linewidth}{!}{
  \begin{tabular}{c|cccc}
    \toprule
    Blocks  & H   & QK    & V     & MLP  \\
    \midrule
    DeiT    & EMB/64 & 64 & 64 & EMB$\times$4 \\
    \midrule
    ReViT     & $\epsilon \times$EMB/100 & $\epsilon \times$EMB/20 & 64 & $\epsilon \times$EMB$\times$3  \\
    \bottomrule
  \end{tabular}
  }
\vspace{-5pt}
\end{table}
For a 12-layer vision transformer model, we use $\epsilon = 2$ for block 4-9, and use $\epsilon = 1$ for other blocks. H is rounded to the nearest even number, and QK rounded to the nearest number divisible by 8 to satisfy Ampere GPUs requirements. 


\textbf{Comparison with DeiT.} 
To verify that our parameter redistribution is beneficial, we train all pairs of DeiT and ReViT models from scratch on the ImageNet-1K benchmark with the same objective and hyperparameters, as specified in Appendix~\ref{ap:hyper_he}. As shown in~\cref{tab:nvit}, ReViT achieve higher accuracy than DeiT with similar FLOPs and lower latency. Specifically, ReViT-S and ReViT-T achieve a Top-1 accuracy gain of 0.21\% and 1.36\%, respectively, over their DeiT counterparts. We also show ReViT rule can work out of the box on SWIN models in Appendix~\ref{ap:swin}.

\begin{table}[htb]
\vspace{-5pt}
  \caption{\small \textbf{Comparing ReViT models with DeiT models.} All compression ratios and speedups are computed with respected to that of the DeiT-Base model. DeiT accuracy marked with * indicates the train-from-scratch accuracy we achieve from the DeiT GitHub repo$^2$ using default hyperparameters$^3$. \textbf{All pairs of models are trained with the same hyperparameters}} 
  \label{tab:nvit}
  \centering
  \resizebox{0.9\linewidth}{!}{
  \begingroup
    \setlength{\tabcolsep}{3pt}
  \begin{tabular}{c|ccccc}
    \toprule
    Model & EMB   & \#Para ($\times$)   & \#FLOPs ($\times$) & Speedup & Accuracy     \\
    \midrule
    DeiT-S   & 384 & 22M (3.94)   & 4.6G (3.82)  & 2.29$\times$ & \ \ 81.01\%*    \\
    \rowcolor{lgreen}
    \textbf{ReViT-S}   & 384 & 23M (3.82)   & 4.7G (3.75)  & 2.31$\times$ & \textbf{81.22\%}    \\
    \midrule
    DeiT-T   & 192 & 5.6M (15.28)  & 1.2G (14.01) & 4.39$\times$ & \ \ 72.84\%*    \\
    \rowcolor{lgreen}
    \textbf{ReViT-T}   & 176 & 5.9M (14.64)  & 1.3G (13.69) & 4.75$\times$ & \textbf{74.20\%}    \\
    \bottomrule
  \end{tabular}
  \endgroup
  }
  \vspace{-10pt}
\end{table}
\footnotetext[2]{\url{https://github.com/facebookresearch/deit}.}
\footnotetext[3]{As in Table 9 of~\cite{touvron2021training}.}

\section{Conclusions}

This work proposes a latency-aware global pruning framework that provides significant lossless compression on DeiT-Base model, facilitating the finding of parameter redistribution for better efficiency-accuracy tradeoff in vision transformers.
We hope this work opens up a new way to better understand the contribution of different components in the ViT architecture, and inspires more efficient ViT models.

{\small
\bibliographystyle{ieee_fullname}
\bibliography{reference}
}

\clearpage
\appendix

\section{Pruning and training details}

\subsection{Training hyperparameters}
\label{ap:hyper}

In our experiments, we use the same data preprocessing, data augmentation, optimizer setup, and learning rate scheduling scheme as mentioned in Table 9 of the DeiT paper~\cite{touvron2021training}, unless otherwise mentioned in the following sections.

\subsubsection{Pruning and finetuning}
\label{ap:hyper_ft}
For pruning and finetuning we use the training objective $\mathcal{L} = \alpha \mathcal{L}_{\text{full}} + \mathcal{L}_{\text{CNN}}$ to update the model. We set the balancing factor 
$\alpha = 1\cdot 10^{5}$ 
and full model distillation temperature $\tau=20$. For our results reported in~\cref{tab:asvit} without CNN distillation, we set $\tau=3$ for the full model distillation objective. 
The pruning process is performed starting from the pretrained DeiT-Base model, with a fixed learning rate of $0.0002\times\frac{\text{batchsize}}{512}$. We perform the pruning experiments on the cluster of four NVIDIA V100 32G GPUs, with a batch size of 128 on each GPU. We prune the model continuously until a targeted latency is reached, which is discussed in detail in Appendix~\ref{ap:config}. Followed by the iterative pruning we remove the pruned away dimensions of the pruned model to turn it into a small and dense model, and continue to finetune the small model to further recover accuracy. Entire finetuning is performed for 300 epochs with an initial learning rate of $0.0002\times\frac{\text{batchsize}}{512}$, cosine learning rate scheduling and no learning rate warm up. The finetuning is performed on a cluster of 32 NVIDIA V100 32G GPUs, with a batch size of 144 on each GPU.

\subsubsection{Downstream tasks transfer learning}
\label{ap:downstream}

\begin{table}[htb]
  \caption{Datasets used for downstream task experiments.}
  \label{tab:data}
  \resizebox{\linewidth}{!}{%
  \centering
  \begin{tabular}{c|ccc}
    \toprule
    Dataset & Train size   & Test size   & \# Classes    \\
    \midrule
    CIFAR-10~\cite{krizhevsky2009learning} & 50,000 & 10,000 & 10 \\
    CIFAR-100~\cite{krizhevsky2009learning} & 50,000 & 10,000 & 100 \\
    iNaturalist 2018~\cite{van2018inaturalist} & 437,513 & 24,426 & 8,142 \\
    iNaturalist 2019~\cite{van2018inaturalist} & 265,240 & 3,003 & 1,010 \\
    \bottomrule
  \end{tabular}
  }
\end{table}

The details of the classification datasets used for our downstream task transfer learning experiments are provided in~\cref{tab:data}.
Similar to the experiment setting of DeiT~\cite{touvron2021training}, for downstream task experiments we rescale all the images to $224\times224$ to ensure we have the same augmentation as the ImageNet training. All models are trained for 300 epochs with a initial learning rate of $0.0005\times\frac{\text{batchsize}}{512}$, cosine learning rate scheduling and 5 epochs of learning rate warm up. We use batch size 512 for CIFAR-10 and CIFAR-100 models, and batch size 1024 for iNaturalist models.

For Semantic Segmentation, previous work SETR~\cite{SETR} provides an effective downstream model architecture and training pipeline to use ViT models as the backbone model of semantic segmentation tasks~\footnote{Code publicly available at \url{https://github.com/fudan-zvg/SETR}.}.
In our experiments we substitute the backbone model with the DeiT/NViT models pretrained on ImageNet. We keep all other downstream architectures and training configurations unchanged. We evaluate the models on the Cityscape dataset~\cite{Cordts2016Cityscapes} and the ADE20K dataset~\cite{zhou2017scene}. For the Cityscape dataset, we follow the ``SETR\_Naive\_DeiT\_768x768\_40k\_cityscapes\_bs\_8'' configuration and train on 4 GPUs. For the ADE20K dataset, we follow the ``SETR\_PUP\_DeiT\_512x512\_160k\_ade20k \_bs\_16'' configuration and train on 2 GPUs.

\subsubsection{ReViT experiments}
\label{ap:hyper_he}

For the experiments on ReViT models we use the CNN hard distillation objective as in~\cref{equ:CNN} as the training objective for all the models. We train Each pair of comparable DeiT and ReViT models with the same set of hyperparameters. 
In all experiments, we train the model from scratch for 300 epochs with an initial learning rate of $0.0005\times\frac{\text{batchsize}}{512}$, cosine learning rate scheduling and 5 epochs of learning rate warm up. The models are trained on a cluster of 16 V100 32G GPUs, with a batch size of 48 on each GPU for base models and a batch size of 144 on each GPU for small and tiny models.

\subsection{Pruning configuration}
\label{ap:config}

\begin{table*}[htb]
  \caption{Pruning configurations and remained dimensions for models reported in Table~\ref{tab:full_compress}. The reported dimensions are averaged across all the blocks. }
  \label{tab:prune_config}
  \centering
  \begin{tabular}{c|cccccccc}
    \toprule
     & & & \multicolumn{5}{c}{Avg. dim remained} \\
    \cmidrule{4-8} 
    Model & Target speedup   & Pruning steps   & EMB & H & QK & V & MLP     \\
    \midrule
    DeiT-B & N/A & 0 & 768 & 12 & 64 & 64 & 3072 \\
    \midrule
    NViT-B & 1.85$\times$ & 480 & 496 & 8.00 & 35.33 & 58.67 & 1917.3 \\
    NViT-H & 2.00$\times$ & 524 & 480 & 7.33 & 32.67 & 56.67 & 1816.0 \\
    NViT-S & 2.56$\times$ & 642 & 400 & 5.83 & 24.00 & 47.33 & 1557.3 \\
    NViT-T & 5.26$\times$ & 908 & 224 & 3.17 & 14.67 & 34.00 & 930.67 \\
    \bottomrule
  \end{tabular}
\end{table*}

We use DeiT-Base model with CNN distillation as the starting point of our pruning process, whose pretrained model is available at \url{https://dl.fbaipublicfiles.com/deit/deit_base_distilled_patch16_224-df68dfff.pth}.
We prune the model in an iterative manner: We compute the moving average of the latency-aware importance score $\mathcal{I}^L_{\mathcal{S}}$ for all unpruned dimension groups in each training step of the pruned model. 
Every 100 steps, we remove a group of dimensions that has the minimum total importance. 
Removed dimensions will never be reactivated.
We prune EMB and MLP in a group size of 16, QK and V in a group size of 8, and H in a group size of 2, so that the input and output dimensions of all the linear projection operations in the model can be divided by 16, thus satisfying the dimension requirement of the Ampere GPU.


 The pruning process will terminate once the estimated latency of the model reaches a targeted speedup ratio over that of the DeiT-base model. The pruned model will then be converted into a small dense model and finetuned to further restore the accuracy. The pseudo code of our pruning algorithm is provided in~\cref{alg}

\begin{algorithm}
\footnotesize
	\caption{Hessian-based latency-aware pruning.} 
	\label{alg}
	\begin{algorithmic}[1]
	    \State \texttt{\# Initialization and preparation}
		\State Load pretrained DeiT-B model
		\State Profile latency lookup table as in Appedix~\ref{ap:lat}
	    \State \texttt{\# Iterative pruning}
		\While {Estimated latency > target}
			\For {$(X,Y)$ in Train\_Loader}
			    \For {All prunable structural group $\mathcal{S}$}
			        \State Compute $\mathcal{I}_{\mathcal{S}}$ with $(X,Y)$ following Equation~(\ref{equ:imp_sim})
			        \State Estimate latency improvement for pruning $\mathcal{S}$
			        \State Compute $\mathcal{I}^L_{\mathcal{S}}$ following Equation~(\ref{equ:imp})
			    \EndFor
			    \State Remove the structural group with $\min_{\mathcal{S}}\mathcal{I}^L_{\mathcal{S}}$
			    \State Estimate pruned model latency 
			    \State Gradient descent on remaining weights
			\EndFor
		\EndWhile
		\State \texttt{\# Finetuning}
		\State Finetune pruned model
	\end{algorithmic} 
\end{algorithm}

\cref{tab:prune_config} reports the target speedup ratio we use to achieve NViT-B, NViT-H, NViT-S and NViT-T architectures reported in~\cref{tab:full_compress}. The resulted number of pruning steps and the averaged dimension of EMB, H, QK, V and MLP among all the transformer blocks are also provided.

\subsection{Latency lookup table profiling detail}
\label{ap:lat}

We use a latency lookup table to efficiently evaluate the latency of the pruned model given all its EMB, H, QK, V  and MLP dimensions. 
We initialize the lookup table by profiling the latency of a single vision transformer block on a V100 GPU with batch size 576. 
We evaluate the latency through a grid of:
\begin{itemize}
    \item EMB: 0, 256, 512, 768 (latency assigned as 0 at zero EMB);
    \item H: 1, 3, 6, 9, 12;
    \item QK: 1, 16, 32, 48, 64;
    \item V: 1, 16, 32, 48, 64;
    \item MLP: 1, and 128 to 3072 with interval 128; 
\end{itemize}
resulting into 9375 configurations in total. We run each configuration for 100 times and record the median latency value in the lookup table.  
For a block with arbitrary dimensions, its latency is estimated via a linear interpolation of the lookup table, which we implement with the \textit{RegularGridInterpolator} function from \textit{SciPy}~\cite{weiser1988note,2020SciPy-NMeth}. 
The estimated latency of the entire model is computed as the sum of the estimated latency of all the blocks, while omitting the latency of the first projection layer and the final classification FC layer.

\begin{figure}[htb]
\centering
\captionsetup{width=0.9\linewidth}
\includegraphics[width=0.7\linewidth]{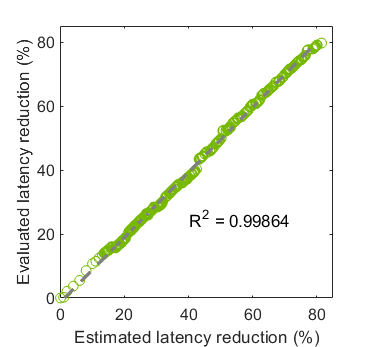}
\caption{Estimated latency from the lookup table vs. evaluated latency on V100 GPU with batch size 256. Reduction ratio computed with respect to the latency of the full model.}
\label{fig:latlut}
\end{figure}

To show the usefulness of the lookup table, we compare the estimated and evaluated latency of different model architectures in~\cref{fig:latlut}. Each point represent the model achieved from a pruning step towards \textit{NViT-T} configuration (See Appendix~\ref{ap:config}). The estimated latency and evaluated latency of ViT demonstrate \textbf{strong} linear relationship throughout the pruning process, with $\mathbf{\mathcal{R}^2 = 0.99864}$. This enables us to accurately estimate the latency improvement brought by removing each group of dimensions, and to use the estimated speedup of the pruned model as the stopping criteria of the pruning process.

\section{Additional ablation studies}

\subsection{Training objective}
\label{ap:obj}

As discussed in~\cref{ssec:objective}, we propose to use a combination of full model distillation and CNN hard distillation as the final objective of our pruning and finetuning process. Here we ablate the validity of this choice and compare the finetuning performance achieved with removing one or both distillation loss from the objective. Specifically, we consider the following 4 objectives:
\begin{itemize}
    \item Proposed objective: $\mathcal{L} = \alpha \mathcal{L}_{\text{full}} + \mathcal{L}_{\text{CNN}}$;
    \item CNN distillation only: $\mathcal{L}_{\text{CNN}}$ as in~\cref{equ:CNN};
    \item Full model distillation with cross-entropy: $\mathcal{L}_{\text{full}} + \mathcal{L}_{\text{CE}}\Big(\Psi\left(\frac{z^s_c+z^s_d}{2}\right),Y\Big)$;
    \item Cross-entropy only: $\mathcal{L}_{\text{CE}}\Big(\Psi\left(\frac{z^s_c+z^s_d}{2}\right),Y\Big)$.
\end{itemize}

We use each of the 4 objectives to finetune the pruned model achieved with NViT-T configuration, and report the final Top-1 accuracy in~\cref{tab:objective}. The finetuning is performed for 50 epochs, with all other hyperparameters set the same as described in Appendix~\ref{ap:hyper_ft}. The proposed objective achieves the best accuracy.

\begin{table}[htb]
  \caption{NVP-T model finetuning accuracy with different objectives.}
  \label{tab:objective}
  \resizebox{\linewidth}{!}{
  \centering
  \begin{tabular}{c|cccc}
    \toprule
    Objective & Proposed   & CNN   & Full model & CE only     \\
    \midrule
    Top-1 Acc.& \textbf{73.55} & 73.40 & 72.62 & 72.36 \\
    \bottomrule
  \end{tabular}
  }
\end{table}

\subsection{Pruning individual components}
\label{ap:ind}

In this section we show the result of pruning EMB, MLP, QK and V component individually. The pruning procedure and objective are almost the same as described in~\cref{ssec:procedure}, other than here we only enable the importance computation and neuron removal on a single component.
The pruning interval of EMB, MLP, QK and V are set to 1000, 50, 200 and 200 respectively, in order to allow the model to be updated for similar amount of steps when pruning different components to the same percentage. 
32 neurons are pruned for each pruning step. We stop the pruning process and finetune the model for 50 epochs after the targeted pruned away percentage is reached. 

The compression rate and accuracy achieved by pruning each component are discussed in~\cref{tab:single}. Under similar pruned away ratio, we can observe that pruning EMB leads to the most significant compression on the parameter and FLOPs count, as well as the largest drop in accuracy. This implies that the embedding dimension leads to the most effective exploration on the compression-accuracy tradeoff, which motivates us to use EMB as the key driving factor in analyzing the parameter redistribution in~\cref{ssec:heur}. 

\begin{table}[htb]
  \caption{Iterative pruning single component to targeted percentage.}
  \label{tab:single}
  \centering
  \resizebox{\linewidth}{!}{
  \begin{tabular}{c|c|ccc}
    \toprule
    Component & Pruned away   & Para ($\times$)   & FLOPs ($\times$)  & Top-1 Accuracy  \\
    \midrule
    Base    & 0\%         & 1     & 1     &           \\
    \midrule
    \rowcolor{lgreen}
    EMB     & 50\%        & 1.98  & 1.92  & 79.24     \\
    MLP     & 50\%        & 1.49  & 1.47  & 82.13     \\
    QK      & 50\%        & 1.09  & 1.10  & 82.98     \\
    V       & 50\%        & 1.09  & 1.10  & 82.63     \\
    \midrule
    \rowcolor{lgreen}
    EMB     & 70\%        & 2.95  & 2.77  & 73.15     \\
    MLP     & 75\%        & 1.97  & 1.91  & 80.29     \\
    QK      & 75\%        & 1.14  & 1.16  & 82.64     \\
    V       & 75\%        & 1.14  & 1.16  & 81.51     \\
    \bottomrule
  \end{tabular}
  }
\end{table}

\subsection{Effectiveness of head alignment}
\label{ap:scheme}

\begin{figure}[htb]
\centering
\captionsetup{width=0.9\linewidth}
\includegraphics[width=\linewidth]{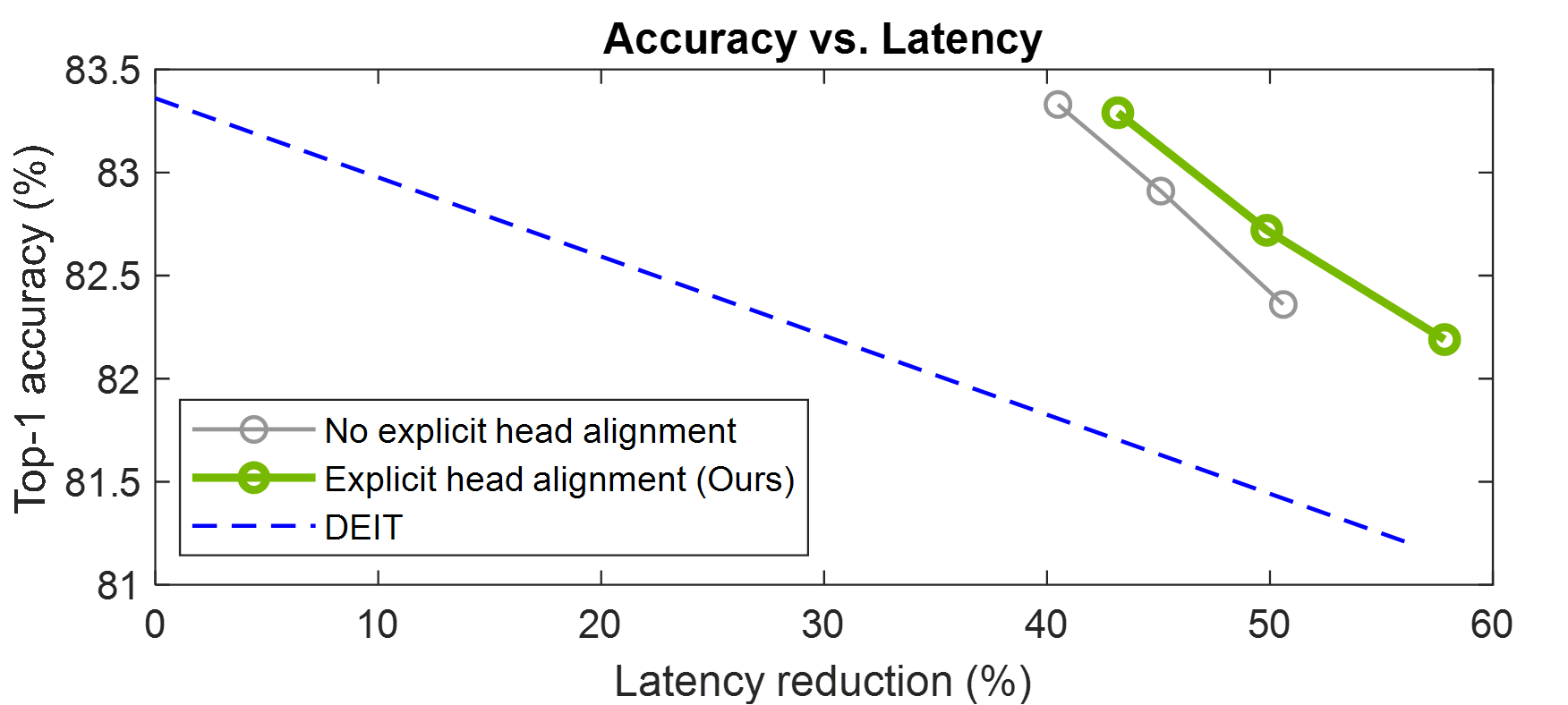}
\caption{Comparing the parameter reduction-accuracy tradeoff and latency reduction-accuracy tradeoff of different pruning schemes. Latency estimated on RTX 2080 GPU. Model size compression rate and latency reduction rate are computed based on that of the DeiT-Base model respectively.}
\label{fig:tradeoff}
\end{figure}

We also illustrate the benefit of head alignment, where we explicitly single out the head dimension and align the dimensions of each head in structure pruning. We show the tradeoff curve between latency reduction and the accuracy achieved with or without explicit head alignment in~\cref{fig:tradeoff}.
For models pruned without head alignment, we estimate their latency as if all heads are padded with zeros to have the same QK and V dimensions during inference.
Under the same latency target, the accuracy achieved with our proposed head-aligned pruning scheme consistently outperforms that of without head alignment, with up to 0.3\% accuracy gain. 

\subsection{Effectiveness of Hessian importance score}
\label{ap:mag}

In our pruning method we claim that utilizing a Hessian-based importance score is the key factor to allow global structural pruning in the ViT models. Here we perform perform an ablation study on pruning with the magnitude-based criteria, where the group with the smallest L2 norm will be pruned in each step. We prune the model to match the latency of DeiT-S, and compare with our NViT-S performance. All the other hyperparameters are set the same. Results are shown in~\cref{tab:mag}.
\begin{table}[htb]
  \caption{Comparing magnitude-based pruning vs proposed NVP. The pruned model accuracy before finetuning is reported.}
  \label{tab:mag}
  \resizebox{\linewidth}{!}{
  \centering
  \begin{tabular}{c|cccc}
    \toprule
    Method & Pruning steps   & Para ($\times$)   & FLOPs ($\times$)  & Top-1 Accuracy  \\
    \midrule
    Magnitude    & 968        & 4.14  & 4.26  & 33.79          \\
    \rowcolor{lgreen}
    NViT-S        & 642        & 4.18  & 4.24  & 76.59     \\
    \bottomrule
  \end{tabular}
  }
\end{table}
It can be seen that magnitude-based pruning struggles to reach the latency target with a larger number of steps, while the pruned model accuracy is much worse. Looking at the remained dimension of the magnitude-based pruning unveils that most of the structural components are either unpruned or all pruned away, which infers magnitude-based criteria is incomparable across different structural components and different layers, thus unsuitable for global pruning. 

\subsection{Correlation between Hessian importance score and real loss difference}
\label{ap:corr}

\begin{figure}[h]
\centering
\captionsetup{width=0.9\linewidth}
\includegraphics[width=\linewidth]{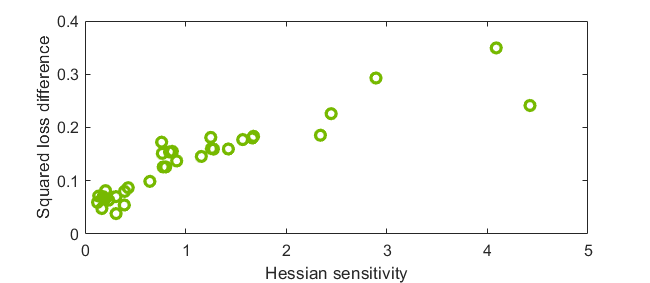}
\caption{Hessian importance score vs. squared loss difference.}
\label{fig:sensitivity}
\end{figure}
In this section we verify the theoretical result derived in~\cref{sssec:Hessian}, on estimating the loss difference induced by pruning with the proposed Hessian importance score. We evaluate the squared model loss increase for performing a single structural pruning step on different structural components of the DeiT-B model, and plot it with the corresponding importance score computed for the pruned structure following the derivation in~\cref{equ:imp_sim}. All the loss differences and Hessian importance score are estimated on the same batch of 64 training images. As shown in~\cref{fig:sensitivity}, we observe strong positive correlation between the estimated sensitivity and the real loss difference.

\subsection{Effectiveness of latency-aware regularization}
\label{ap:latreg}

\begin{table*}[htb]
  \caption{Comparing pruning results with ($\eta=$5e-4) or without ($\eta=0$) latency-aware regularization. The pruned model accuracy before finetuning is reported. The reported dimensions are averaged across all the blocks. }
  \label{tab:latreg}
  \centering
  \begin{tabular}{c|ccccccccccc}
    \toprule
     & & & & & \multicolumn{5}{c}{Avg. dim remained} \\
    \cmidrule{6-10} 
    $\eta$ & Pruning steps & Para ($\times$)   & FLOPs ($\times$)  & Acc.   & EMB & H & QK & V & MLP     \\
    \midrule
    0. & 657 & 4.11 & 4.17 & 74.80 & 416 & 5.7 & 25.3 & 49.3 & 1510.7 \\
    \rowcolor{lgreen}
    5e-4 (NViT-S) & 642 & 4.18 & 4.24 & 76.59 & 400 & 5.8 & 24.0 & 47.3 & 1557.3 \\
    \bottomrule
  \end{tabular}
\end{table*}

In~\cref{tab:latreg} we show the result of pruning without latency regularization, i.e. set $\eta=0$ in the importance score formulated in~\cref{equ:imp}, and compare with our NViT results. Both models are pruned to match DeiT-S latency. 
We can see from the result that pruning with latency-aware regularization can help reaching the target latency quicker, while achieving higher accuracy under the latency budget. To better understand the difference in the achieved architecture, we also show the average dimension across all the blocks after pruning.
It can be seen that model pruned with latency regularization tends to have more dimensions on MLP and less on MSA (QK and V), which is in line with our observation made in~\cref{ssec:heur} on designing more efficient ViT architecture, where reducing dimensions related to the attention (H, QK, V) while increasing MLP dimension may lead to more accurate model under similar latency.

\subsection{Performance on low-end GPUs}
\label{ap:nano}

As one of the main motivation for pruning is to enable model deployment on low-end devices with cheaper cost and lower energy consumption. To this end we further examine the latency of running the pruned NViT models on NVIDIA Jetson NANO, a commonly used low-end GPU for embedded system. Here we utilize a batch size of 64 for ImageNet inference. 

For base model, we note that DeiT-B cannot fit into the memory of the device, preventing it from being compiled onto the NANO device. Whereas our pruned NViT-B model can run with a decent speed, reaching $83.3\%$ Top-1 acc. NViT-T matches the speed of DeiT-T, and the speedup over NViT-B is consistent to our measurement on V100 reported in~\cref{tab:full_compress} ($2.8\times$ on NANO vs. 2.7$\times$ on V100). This further demonstrates that for low-end devices NViT enables the originally prohibitive high-performance model to run, while the speedup achieved on high-end devices can be retained.

\section{Additional parameter redistribution analysis}

\subsection{Attention head diversity}
\label{ap:intuition}

\begin{figure*}[htb]
\centering
\captionsetup{width=0.9\linewidth}
\includegraphics[width=0.9\linewidth]{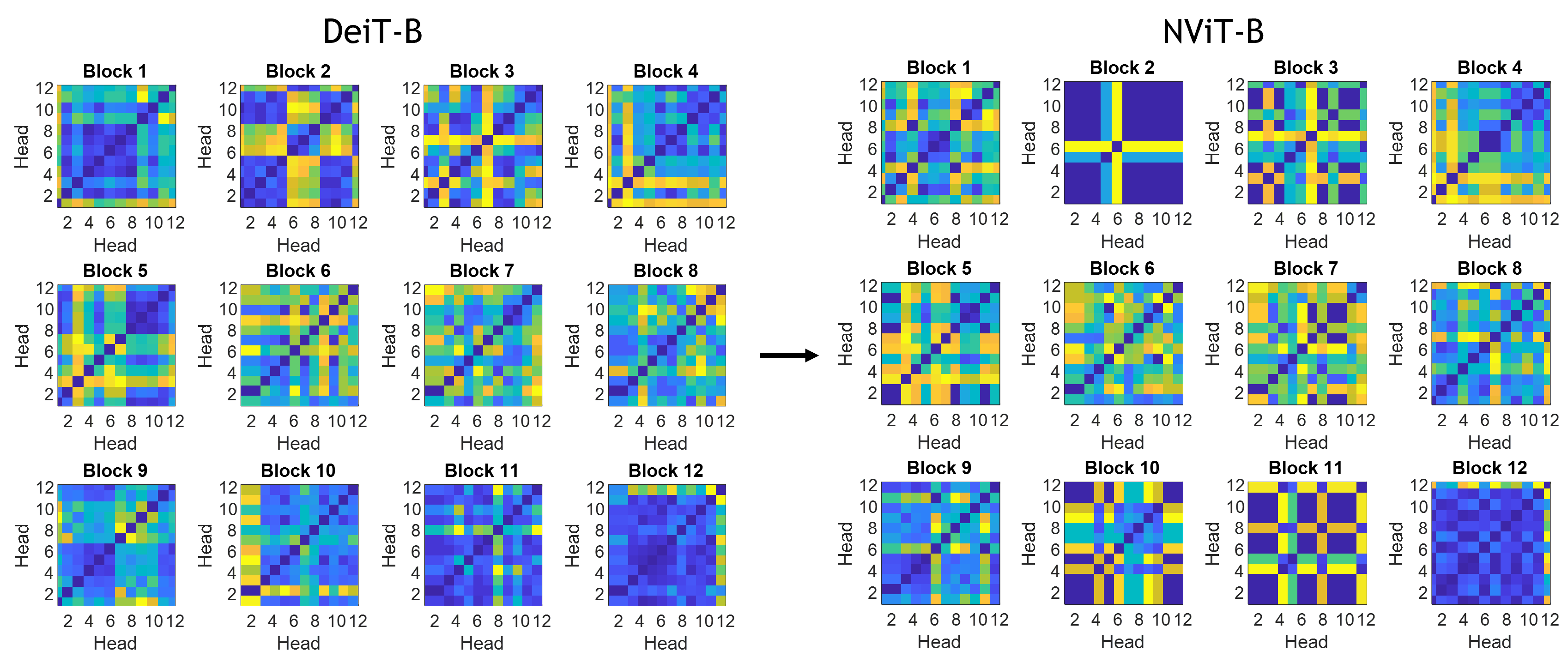}
\caption{Pair-wise cosine distance between all heads' attention score in each transformer block. Blue indicates a smaller distance while yellow indicates a larger one. The dark blue blocks in NViT-B figures corresponds to the heads being pruned away, which have all-zero attention scores thus zero cosine distance in between.}
\label{fig:attn}
\end{figure*}

As we observe in~\cref{fig:heur} and mentioned in~\cref{ssec:heur}, the pruned models tend to preserve \textit{more} dimensions in the transformer blocks towards the \textit{middle} layers, while having \textit{less} dimensions towards the \textit{two ends} of the model. Here We explore an intuitive analysis on why this trend occurs by observing the diversity of features captured in each transformer blocks. 
Given the attention computation serves important functionality in ViT models, here we use the diversity of the attention score learned by each head as an example. Specifically, we take a random batch of 100 ImageNet validation set images, pass them through the pretrained DeiT-Base model and our NViT-B model, and record the averaged attention score $softmax\left(\frac{q_h k_h^T}{\sqrt{d_h}}\right)$ of all the images computed in each head $h$. We then compute the pair-wise cosine distance of the attention score from each head as a measure of diversity, and visualize the results in~\cref{fig:attn}.

In DeiT-B model, we can observe that in earlier blocks like block 2 and later blocks like block 11, there are clear patches of darker blue indicating a group of heads having attention scores similar to each other. While for blocks in the middle such as block 5-8, almost all pairs of heads appear to be fairly diverse. Such difference in diversity leads to different behavior in the pruning process, where less heads are preserved in earlier and later blocks while more are preserved in the middle. Note that all remaining heads in NViT-B model appears to be diverse with each other, showing a more efficient utilization of the model capacity.
Interestingly, this less-more-less trend of dimensional change across different transformer is not observed in previous works compressing BERT model for NLP tasks~\cite{voita2019analyzing,michel2019sixteen,mao2021tprune}. The learning dynamic of ViT model leading to this trend is worth investigating in the future work. 

\subsection{Parameter redistribution on SWIN}
\label{ap:swin}
We have shown the effectiveness of the proposed pruning method on pruning SWIN-Transformer stages. In this section,  we examine the effectiveness of the discovered parameter redistribution rule of DeiT on the Swin-Transformer model. Though SWIN follows a multi-stage design that is different from DeiT, within each stage all the transformer blocks have the same dimension, which gives us the potential of exploring better dimension redistribution rules. 
Here we take SWIN-T model, with 2-2-6-2 transformer blocks in stage 0-3 respectively. As the redistribution rule treats the first/last block and intermediate blocks differently, the rule mainly takes effect on stage 2 with 6 blocks. 
The parameter redistribution is performed following exactly the same ReViT rule as reported in~\cref{tab:heur_sim}. 
Specifically, the dimensions of each transformer block in the redistributed SWIN-ReViT-T is reported in~\cref{tab:SWIN-NViT}.

\begin{table}[h]
  \caption{Redistributed SWIN-ReViT-T model Stage-2 dimensions.}
  \label{tab:SWIN-NViT}
  \resizebox{\linewidth}{!}{
  \centering
  \begin{tabular}{c|cccccc}
    \toprule
    Block & 1 & 2 & 3 & 4 & 5 & 6 \\
    \midrule
    EMB & 384 & 384 & 384 & 384 & 384 & 384 \\
    Head & 10 & 4 & 8 & 8 & 4 & 10 \\
    QK/Head & 32 & 16 & 32 & 32 & 16 & 32 \\
    V/Head & 64 & 64 & 64 & 64 & 64 & 64 \\
    MLP & 1152 & 1152 & 2304 & 2304 & 1152 & 1152 \\
    \bottomrule
  \end{tabular}%
  }
\end{table}

We train the SWIN-ReViT-T model on ImageNet following the same training scheme described in the official GitHub repo~\footnote{\url{https://github.com/microsoft/Swin-Transformer}}. The model statistics and training performance of the resulted SWIN-ReViT-T is compared with the original SWIN-T in~\cref{tab:SWIN-NViT-result}.
\begin{table}[htb]
  \caption{Comparing the efficiency and accuracy of SWIN-ReViT-T vs. SWIN-T on ImageNet. The throughput is evaluated with a single TITAN RTX GPU.}
  \label{tab:SWIN-NViT-result}
  \resizebox{\linewidth}{!}{
  \centering
  \begin{tabular}{c|cccc}
    \toprule
    Model & Parameters & FLOPs & Throughput & Top-1 Accuracy \\
    \midrule
    SWIN-T & 29M & 4.5G & 546.37 img/s & 81.3\% \\
    \rowcolor{lgreen}
    \textbf{SWIN-ReViT-T} & \textbf{28M} & \textbf{4.4G} & \textbf{574.25} img/s & 81.3\% \\
    \bottomrule
  \end{tabular}%
  }
\end{table}

The redistributed SWIN-ReViT-T model achieves the same Top-1 accuracy as the original model with 1.1x speedup. This indicates that the redistribution rule derived on DeiT can also be transferred to other ViT variants to achieve efficiency improvements. 

\subsection{The significance of ReViT-S performance gain}
\label{ap:nvit_sig}

\begin{table}[htb]
  \caption{Repeated experiments of ReViT-S and DeiT-S training.}
  \label{tab:repeat}
  \resizebox{\linewidth}{!}{
  \centering
  \begin{tabular}{c|ccccc|cc}
    \toprule
    Model & Ckpt 1  & Ckpt 2  & Ckpt 3  & Ckpt 4  & Ckpt 5  & Mean  & STD  \\
    \midrule
    DeiT-S & 80.96 & 80.93 & 80.95 & 81.01 & 80.92 & 80.954 & 0.035 \\ 
    \rowcolor{lgreen}
    \textbf{ReViT-S} & 81.17 & 81.19 & 81.17 & 81.20 & 81.22 & 81.190 & 0.021 \\
    \bottomrule
  \end{tabular}
  }
\end{table}

As we report the accuracy improvement brought by ReViT-S over DeiT-S in~\cref{tab:nvit}, here we verify the significance of this improvement via repeated experiments. Specifically, we report the Top-1 accuracy of 5 checkpoints for training ReViT-S and DeiT-S from scratch on ImageNet in~\cref{tab:repeat}.
Note that the averaged 0.23\% Top-1 accuracy gain of ReViT-S over DeiT-S is 10 times the standard derivation of repeated experiment results, showing the improvement is truly significant.

\end{document}